\documentclass[10pt,twocolumn,letterpaper]{article}

\usepackage{iccv}
\usepackage{times}
\usepackage{epsfig}
\usepackage{graphicx}
\usepackage[tight]{subfigure}
\usepackage{amsmath}
\usepackage{amssymb}

\usepackage[pagebackref=true,breaklinks=true,letterpaper=true,colorlinks,bookmarks=false]{hyperref}

 \iccvfinalcopy 

\def\iccvPaperID{****} 

\ificcvfinal\fi
\begin{document}

\title{The Conditional Analogy GAN: Swapping Fashion Articles on People Images}

\author{Nikolay Jetchev\\
Zalando Research\\
{\tt\small nikolay.jetchev@zalando.de}
\and
Urs Bergmann\\
Zalando Research\\
{\tt\small urs.bergmann@zalando.de}
}

\maketitle

\begin{abstract}
We present a novel method to solve image analogy problems~\cite{Hertzmann2001}: it allows to learn the relation between paired images present in training data, and then generalize and generate images that correspond to the relation, but were never seen in the training set. Therefore, we call the method Conditional Analogy Generative Adversarial Network (CAGAN), as it is based on adversarial training and employs deep convolutional neural networks.
An especially interesting application of that technique is automatic swapping of clothing on fashion model photos. 
Our work has the following contributions. First, the definition of the end-to-end trainable CAGAN architecture, which implicitly learns segmentation masks without expensive supervised labeling data. Second, experimental results show plausible segmentation masks and often convincing swapped images, given the target article.
Finally, we discuss the next steps for that technique: neural network architecture improvements and more advanced applications.
\end{abstract}

\section{Introduction}
\subsection{Image content challenges in the fashion business}
In modern fashion e-commerce, scale and speed is both an opportunity and a challenge. The creation of image content for the huge amount of various articles means that traditional methods (shooting and photographing models) are a bottleneck, as they're slow and expensive.
Suppose that several thousands of articles arrive in a new batch. Shooting them on cloth hangers standalone is relatively easy and cheap, but making photoshoots with professional models is time consuming and expensive. Leveraging available data and reusing images of human models and products would therefore be very useful for a fashion business.

As a related application, consider the ``virtual try-on" problem:
giving a customer the ability to upload her/his picture and see how she/he would look with different fashion products in a sort of ``magic mirror''\footnote{\url{https://www.slideshare.net/metatechnology/magic-mirror-for-fashion-stores}} will give totally new exciting possibilities to customers in the near future. However, this is currently a technical challenge that has not been solved convincingly.

One approach would be to model humans and articles as 3D objects and render them photo-realistically with computer graphics methods. However, the engineering challenges and computational costs (e.g. for high quality 3D scans ) make that approach expensive to build and maintain. Instead, the rich 2D image collections characteristic for fashion companies contain rich structure that can be exploited by modern generative deep learning methods.

By learning a generative model for ``dressing" an image of a human model given a fashion article photo specifying what to wear, we allow a novel image editing method that has great potential for fashion businesses. Our method can enable a lean system capable of creating thousands of images per second by using image data available in typical fashion e-commerce companies, as produced by the standard content production processes. It thus has the potential to allow further scaling in parts of current fashion businesses.

To avoid confusion, in this paper we speak of ``human" instead of image of a human fashion model, and use the word model only for neural network models. 

\subsection{Image-to-image translation}
Many image processing problems can be modeled as image-to-image translation problems (e.g. colorizing black-and-white images, translating from a sketch of a scene to a colorful photography, etc.), where an image is fed as input and the output is the modified image.

Convolutional neural networks (CNN) are powerful deep learning models that can solve such problems, see \cite{zhang2016colorful} for an example. However, usual CNN architectures require prespecified loss functions and such losses may not be suitable for outputting sharp and realistic images, e.g. a Euclidean loss leads to blurry images.

The more recent model class of Generative Adversarial Networks (GANs)~\cite{Goodfellow14} train a model $G$ that learns a data distribution from example data, and a discriminator $D$ that attempts to distinguish generated from training data. These models learn loss functions that adapt to the data, and this makes them perfectly suitable for image-to-image translation tasks where the desired result is to create sharp images that look indistinguishable from the training examples.

The conditional version~\cite{MirzaO14} of GAN (cGAN) learns to generate images as function of conditioning information from a dataset, instead of random noise from a prior, as in standard GANs.
\cite{pix2pix} is an excellent overview of image-to-image translation methods using cGANs.
\cite{texturegan} further improves the type of losses available for image-to-image translation models and allows advanced sketch and texture control for interactive image editing.
The requirement of datasets with paired input and output images as ground truth is sometimes a limitation. For many interesting image translation problems no such datasets with paired data exist. \cite{cyclegan} presents a novel method of unpaired training for such cases: presenting two (unpaired) image domains, and regularizing the generator function, allows to learn a mapping between the image domains without requiring explicitly paired examples. For example, by getting a set of horse images, and a second set of zebra images, the model can accurately swap their textures.

\section{The CAGAN model}
\subsection{Painting humans wearing fashion articles as an image analogy problem}
For the task of painting a given image of a human with a given image of a certain article, one could potentially train supervised learning methods. However, ground truth data corresponding to the desired outcome does not exist, and it is not practical to generate it manually.
Instead, in a typical fashion company, there's rich image content material produced constantly in photo-shoots: millions of photos of humans models $x$ wearing articles of clothing, and also close-up pictures of the articles $y$ without humans. 
CAGAN will use image data of such form, i.e. $\mathcal{D} = \{x_i,y_i \}_{i=1}^N$, where the index $i$ indicates a pair of human and article. These images have a relation -- the image $x_i$ contains a human wearing a certain fashion product on his body, and $y_i$ is another image of the same fashion product shown alone.
 
The mapping -- between the standalone article image $y_i$ and its appearance rendered on a human in $x_i$ -- is typically distorted by occlusion, illumination, 3D rotation and deformation.
We can use data $\mathcal{D}$ to learn this relation, and then, given a new (not in the training set) clothing article $y_*$, create an image of a human body $x_*$ wearing the appropriately rendered article. This is an image analogy problem: find $x_*$ in the same relation to $y_*$ as $x_i$ is to $y_i$ for all training data pairs $i$. Note that our method, in contrast to the standard image analogy method~\cite{Hertzmann2001}, makes use of the whole dataset $\mathcal{D}$ to infer the most plausible relation.

However, generating a completely new photorealistic image containing a human who wears a fashion article image is a complex task. In particular, plausible faces are very hard to generate~\cite{clothnet}. Fortunately, in our case it is beneficial to restrict the problem by staying close to the input image, as described in the next section.

\begin{figure}[tb]
\begin{center}
\includegraphics[width=8cm]{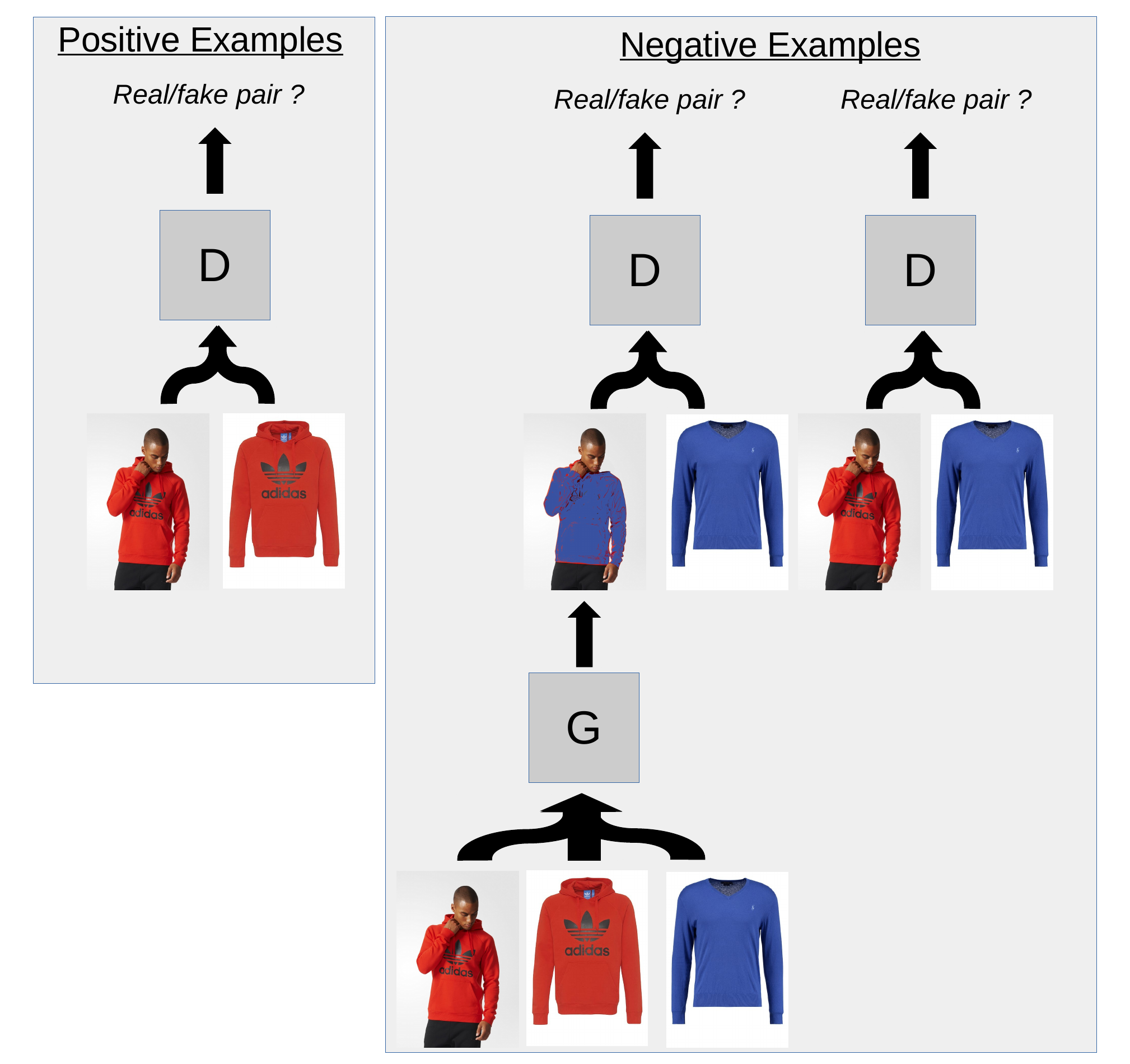}
\end{center}
\caption{The CAGAN: given pairs of humans and clothes, learn to swap clothes and paint realistically looking images.}\vspace*{-0.3cm}
\end{figure}

\subsection{Relaxing the task: swapping clothes on existing humans}
We just need to make sure the fashion article looks well painted, and can reuse an existing human model image, instead of creating a complete output image from scratch. This is easier from a machine learning point of view. In addition it also corresponds to the fashion customization use-case from the introduction, i.e. how will a particular customer look in a piece of clothing?

We want to train an image-to-image translation network that will exchange one piece of clothing $y_i$ with a new one, $y_j$, on a \textit{given} human image $x_i$.
Note there are never examples of   ${x}_i^j$, the modified human image with the swapped fashion item we would like to see. We infer this ${x}_i^j$ indirectly from the data, which allows us to learn the relation $x_i,y_i$, an article properly dressed on a model. The priors and model architecture will force the neural model to augment the human image and paint only some parts of it with the new article, which we consider implicit relation learning. Note that this is more complex than~\cite{cyclegan}, because in our  case the image domains are specified implicitly by the conditioning, rather than given explicitly as two image domains training data. 

Our method incorporates end-to-end learning to side step localization and segmentation challenges and directly predict images having suitable properties: be smooth looking fashion models that wear the specified fashion articles. For this, we need to find where the old article is located, and replace it with the new articles.
We  use as conditioning directly an article stand-alone image, of the type usually available in an online shop. They show the article in some detail, but the transformation to correct for illumination, occlusion, 3D rotation and deformation such that it fits the output is not known. Our method automatically infers an appropriate segmentation map from both the article and human image, and uses it to generate an appropriate looking image consistent with the conditioning.

\subsection{CAGAN model and training loss function}
Training of the CAGAN model involves learning a generator $G$ to generate plausible images which fool a discriminator $D$. The discriminator $D$ needs to answer two questions:
\begin{itemize}
\item does an image $x$ look reasonable, i.e. indistinguishable from the training distribution of human images $\{x_i\}$?
\item does the article $y$ look well-painted on the human model image $x$, i.e. is the relation of $x$ and $y$ consistent with the relations observed in the dataset $\mathcal{D} = \{x_i,y_i \}_{i=1}^N$.
\end{itemize}
The second criteria cannot be modeled directly with an Euclidean loss since we do not have ground truth examples for all the combinations of \textit{any} article $y$ with \textit{any} human $x$ -- the data $\mathcal{D}$ has only a few examples of \textit{specific} humans, each wearing only a few clothes.
 
For training $D$ and $G$ we define a loss which contains several terms, weighted by constants $\gamma_i,\gamma_c$:
\begin{align}
 \min_{G} \max_{D} \mathcal{L}_{cGAN}(G,D)+ \gamma_i \mathcal{L}_{id}(G) +\gamma_c\mathcal{L}_{cyc}(G). \label{eqICGAN}
\end{align}

The most important term is the adversarial loss that involves the generator and the discriminator:
\begin{align}
\mathcal{L}_{cGAN}(G,D) = \mathbb{E}_{x_i,y_i \sim p_{\mathrm{data}}} \sum_{\lambda,\mu} \left[ \log D_{\lambda,\mu}(x_i,y_i) \right]  \nonumber \\
  + \mathbb{E}_{x_i,y_i,y_j \sim p_{\mathrm{data}}}\sum_{\lambda,\mu} \left[ \left(1 - \log D_{\lambda,\mu}(G(x_i,y_i,y_j),y_j) \right) \right] \nonumber \\
    + \mathbb{E}_{x_i,y_{j \neq i} \sim p_{\mathrm{data}}}\sum_{\lambda,\mu} \left[ \left(1 - \log D_{\lambda,\mu}(x_i,y_j) \right) \right], \label{eqGAN} 
\end{align}
where the notation $x_i, y_i, x_{j \neq i} \sim p_{\mathrm{data}}$ means uniformly sampling elements with indices $i$ and $j$ from the dataset $\mathcal{D}$, under the fullfillment of the constraint $j \neq i$.
We marginalise over the indices $\lambda,\mu$ of the spatial dimensions of the output of $D$, because the discriminator does not output a single number per image, but a whole spatial field of classifications. The network $D_{\lambda,\mu}: x,y \mapsto [0,1] $ is a typical discriminative network. Such local consistency discriminator works quite well, see also \cite{pix2pix,SGAN2016,PSGAN2017} where they discuss how discriminating patches from a bigger image and marginalizing over the spatial positions is fast and efficient for many image discrimination tasks.

The terms of loss (\ref{eqGAN}) are very similar to the classical GAN loss of \cite{Goodfellow14} that learn to distinguish true data from generated examples.
However, the last term of $\mathcal{L}_{cGAN}$ defines another type of negative examples that come from the distribution $p_{\mathrm{data}}$ of swapped data: a human $x_i$ and an article $y_j$ different than $y_i$ and ``not'' appearing on the human. The motivation for that term is that we need to find whether the article is really worn by the human, i.e. the discriminator needs to learn if the relation of $x_i$ and $y_j$ is corresponding to the dataset.
\cite{reed2016generative} also finds such negative examples to be beneficial for conditional generative models.

The generator network $G: x_i,y_i,y_j \mapsto [\alpha_i^j, \tilde{x}_i^j] \mapsto x_i^j $ is designed to output a 4 channel output: a 1 channel matting mask $\alpha_i^j$ and a 3 channel color image $\tilde{x}_i^j$. The final generated image $x_i^j$ of a human with swapped clothes is the convex combination of the predicted and original image $x_i^j = \alpha_i^j \tilde{x}_i^j + (1-\alpha_i^j)x_i$.
In order to ensure a range $[0,1]$ for the $\alpha$ mask values, we transform them with the sigmoid function. 
Such prediction of both blending mask and predicted image is similar to \cite{LRGAN}.

\begin{figure}
\centering
\includegraphics[width=7.5cm]{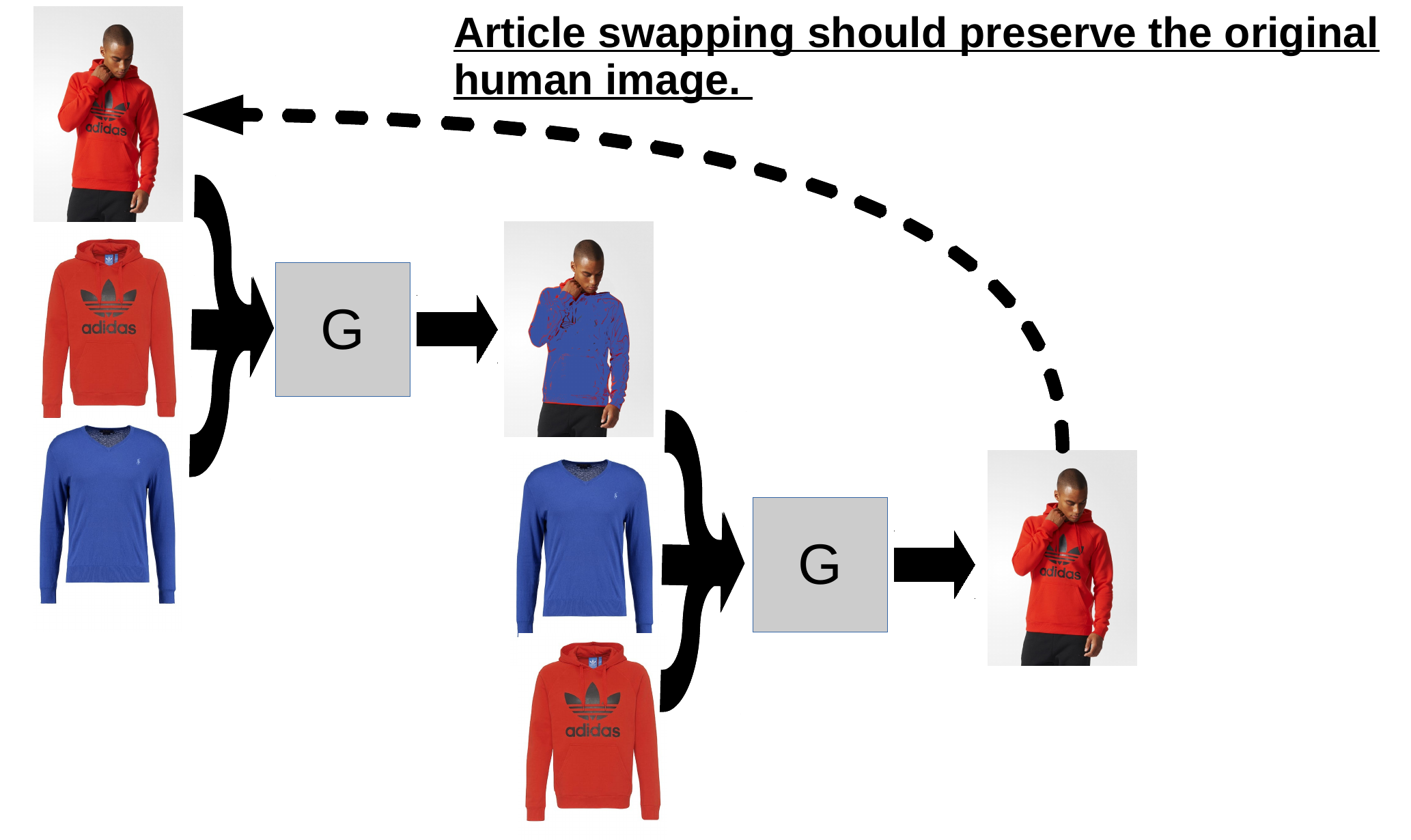} 
\caption{Illustration of the cycle loss: (i) swap article $y_i$ with $y_j$ on image $x_i$ by applying $G$; (ii) swap $y_j$ with $y_i$ and the output of $G$ should be close to the original human model image $x_i$.}\vspace*{-0.3cm}
\label{fig_cycle}
\end{figure}

The term $\mathcal{L}_{id}$ in Eq. (\ref{eqICGAN}) regularizes the outputs of $G$ to change as little as possible from the original human image, since we know we only change one piece of clothing at a time. It has the effect to avoid the model painting parts of the human body that are not relevant for the swapped clothes:
\begin{align}
\mathcal{L}_{id}(G) = \mathbb{E}_{x_i,y_i,y_j \sim p_{\mathrm{data}}}  \|\alpha_i^j\|, \nonumber
\end{align}
where $\|.\|$ is the L1-norm. Our preliminary results indicate that different norms (e.g. total variation) can improve the results further.

We also introduce a cycle loss in equation (\ref{eqICGAN}) that should force consistent results when swapping clothes:

\begin{align}
\mathcal{L}_{cyc}(G) =  \mathbb{E}_{x_i,y_i,y_j \sim p_{\mathrm{data}}} \| x_i - G(G(x_i,y_i,y_j),y_j,y_i)  \|. \nonumber
\end{align}

Figure \ref{fig_cycle} illustrates how the cycle loss is defined, similar to \cite{cyclegan}.
The motivation is that it should make the generator more stable: changing clothes on a human should only swap the relevant articles and leave the other parts of the image unchanged. If the generator $G$ creates an image $x_i^j=G(x_i,y_i,y_j)$ that modifies original image $x_i$ in some regions unrelated to the article $y_j$, then the reverse swapping operation $G(x_i^j,y_j,y_i)$ will generate an image that will be penalized for deviating from $x_i$.


\section{Experiments}
\subsection{Setup}
\begin{figure}
\centering
\includegraphics[width=7.5cm]{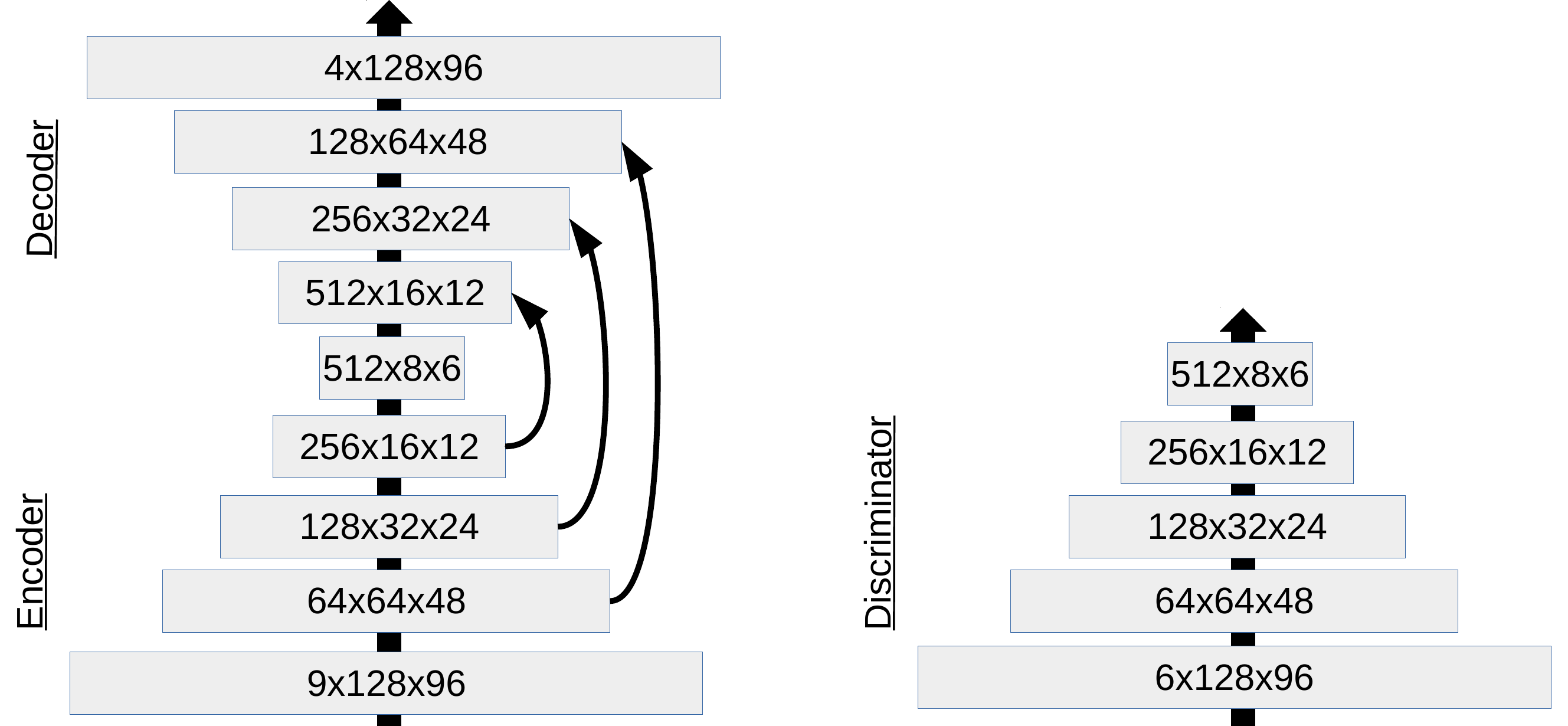} 
\caption{The encoder-decoder generator with skip connections, and the patch discriminator. We display the values "channels \textbf{x} width \textbf{x} height" for illustration of the  image size 128x96 pixel experiments we did. Note that the discriminator outputs a spatial field of classification values, a so-called PatchGAN \cite{pix2pix} approach.}\vspace*{-0.3cm}
\label{unet}
\end{figure}

Our setup used an Nvidia K80 GPU with 8gb of memory, and the code was implemented in Theano.
For training we use ADAM~\cite{KingmaB14} with the settings of~\cite{RadfordMC15} -- learning rate $0.0002$, and minibatch size of 16. We use instance normalisation at all layers except the first and last, as in \cite{pix2pix}, and \textbf{Relu} as activation function in all layers.
Usually several hours of training (10000 gradient steps) were enough to get reasonable results. The strengths of the regularizations for Eq. (\ref{eqICGAN}) are $\gamma_i = 0.1$ and $\gamma_c = 1.0$. In our experience both regularizations are beneficial for the article swapping task, but did not investigate in detail what the optimal values are.

$D$ uses convolutions with stride 2, $G$ has also convolutions with stride 2 and deconvolutions with stride $\frac{1}{2}$. We double the number of channels when the spatial resolution decreases. Figure \ref{unet} shows the spatial sizes and channels when training on 128x96 pixel images.
For $D$ it was enough to use 4 layers with receptive field of only 63x63 pixels, enforcing local consistency when discriminating images.

For $G$ we use the encoder-decoder architecture with skip connections, as in \cite{pix2pix}.
In addition, we use always the last 6 channels of any intermediate layer (in both $G$ and $D$) to store downsampled copies of the inputs $x_i,y_i$. This improves the convergence of the models and the image quality. We suppose in that way information from the conditioning is better preserved in the deep network.

The training data used contained 15000 images of humans (frontal view) and paired upper-body garments  (pullovers and hoodies). The data was given by Zalando SE\footnote{\url{www.zalando.de}}, one of the biggest e-commerce fashion companies. We used the data in \textbf{RGB} color space, and tested both 128x96 and 256x192 pixel resolutions with this data. 

\subsection{Generation results: images of human models with swapped clothes}
\begin{figure}
\centering
\includegraphics[width=8.5cm]{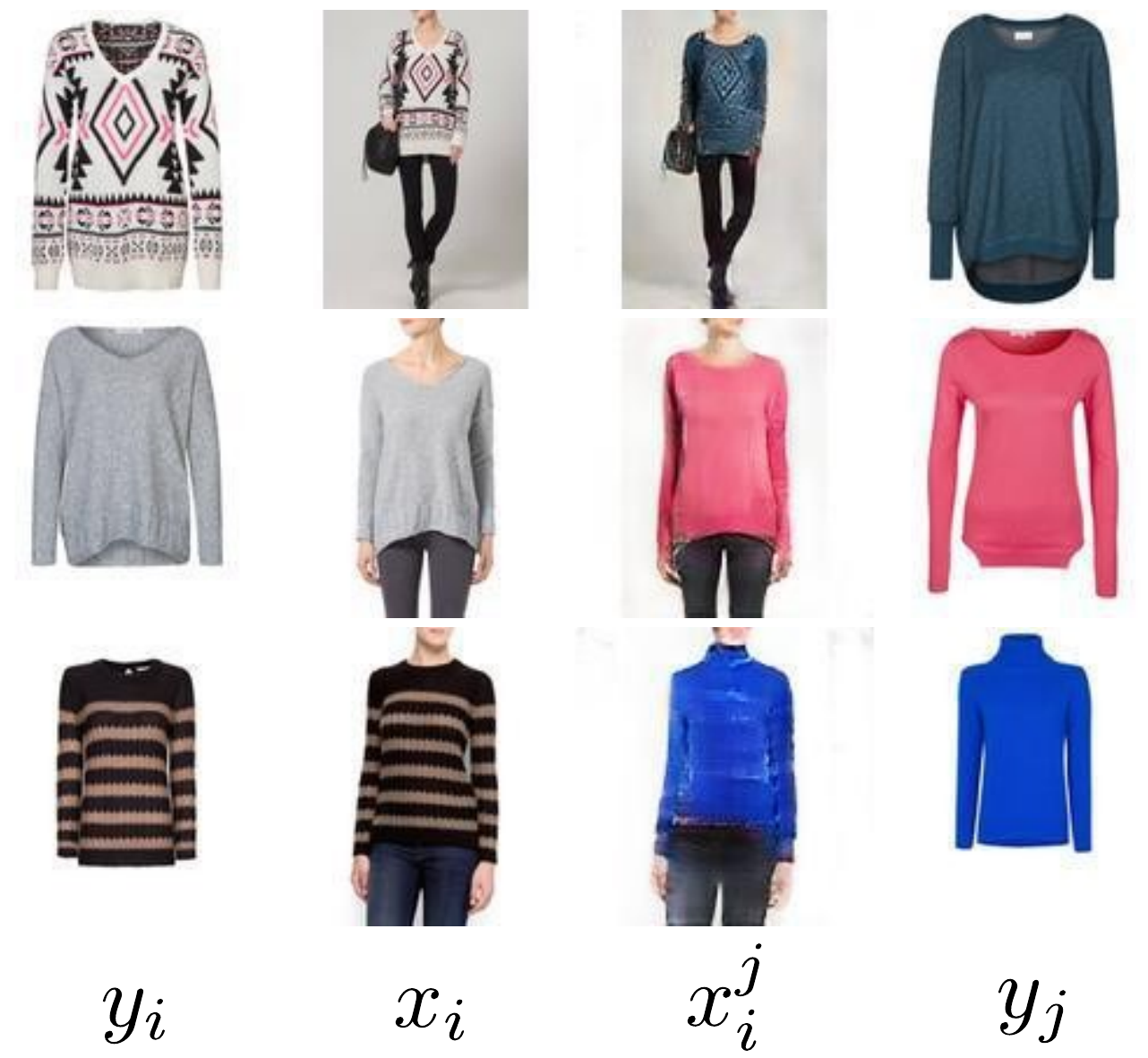} 
\caption{Changing the upper-body garment of a human:  the original $x_i$ wears article $y_i$; we want to paint him wearing article $y_j$; the generated image with that property is $x_i^j$. Results with 128x96 pixels resolution. Note that not-in-place transformations (e.g. the neckline of the pullovers) are also generated by CAGAN.}\vspace*{-0.3cm}
\label{fig_basic}
\end{figure}

In general, our model has the correct behavior: generated images keep the looks of the human image and only swap the relevant article.
Figure \ref{fig_basic} shows our results on 3 randomly chosen models.
In-place color changes are easier than texture and geometric deformations \cite{cyclegan}.

\begin{figure}[htb]
\centering
\includegraphics[width=8cm]{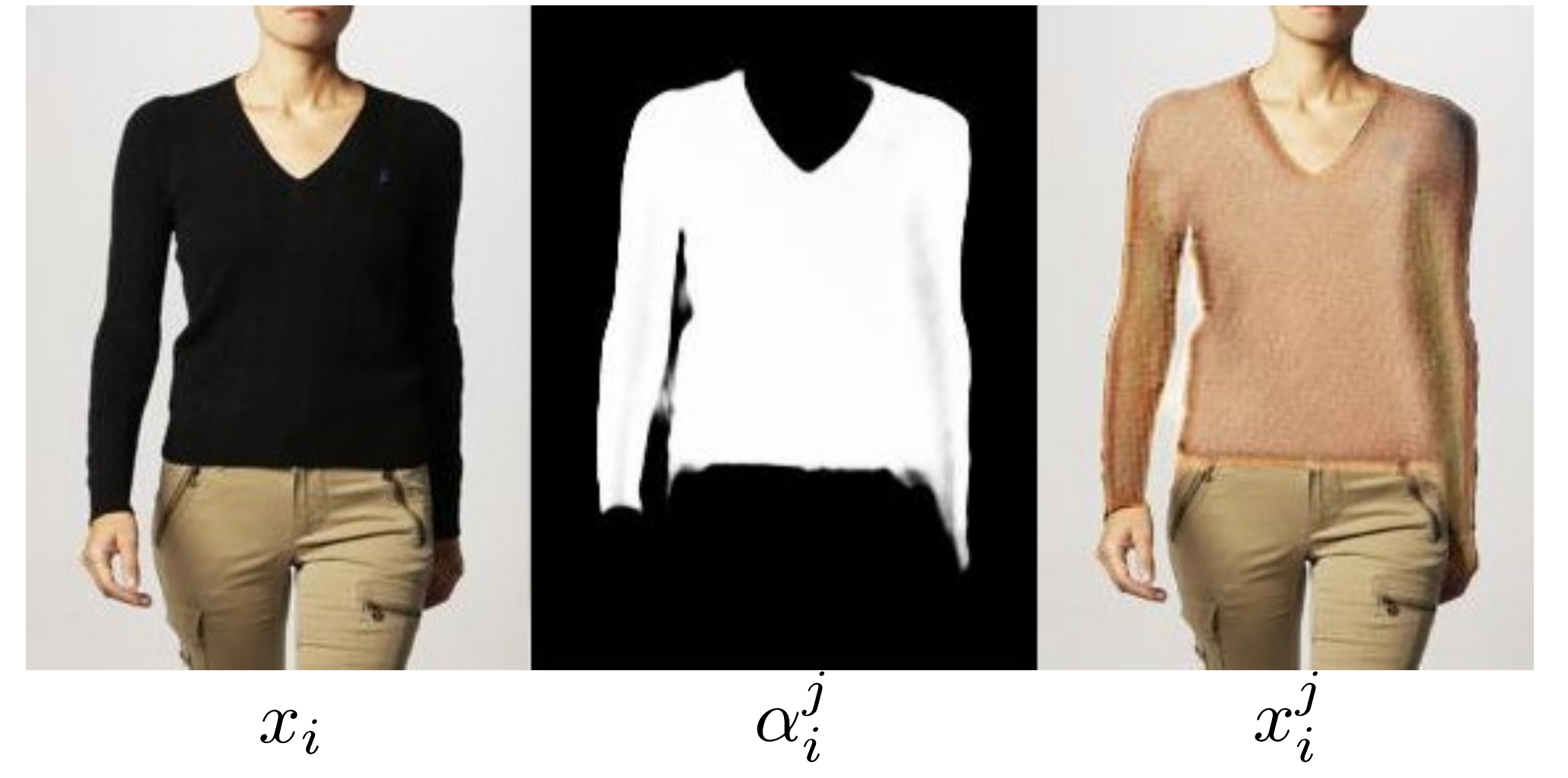} 
\caption{The alpha mask $\alpha_i^j$ implicitly learned by CAGAN is quite accurate for the original image $x_i$. This allows accurate repainting with another fashion article of the human, resulting in $x_i^j$. Results with 256x192 pixels resolution.}\vspace*{-0.25cm}
\label{fig_alpha}
\end{figure}

Figure \ref{fig_alpha} shows results with a higher resolution. We see that the masks $\alpha$ can be of really good quality, which is impressive given that the segmentation is learned implicitly by the CAGAN objective. We note that the images we obtained from Zalando have a consistent background -- street view images would likely be more difficult, since background clutter makes segmentation more complex.

\begin{figure*}
\centering
\subfigure[A set of human models $\{x_i\}_{i=1}^{16}$.]{\fbox{\includegraphics[width=6.9cm]{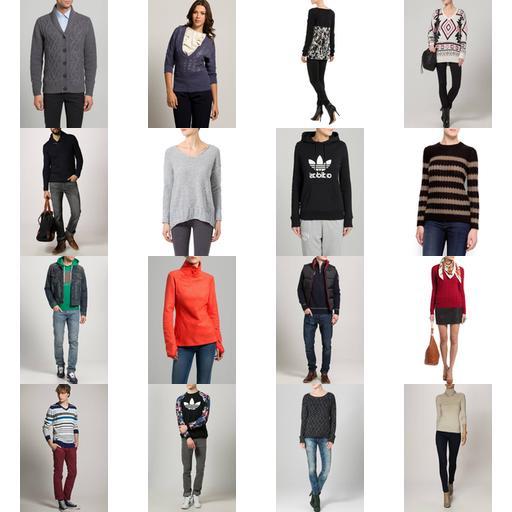}} }\hspace{5mm}
\subfigure[A set of articles $\{y_j\}_{j=1}^{16}$, not present on the images in (a).]{\fbox{\includegraphics[width=6.9cm]{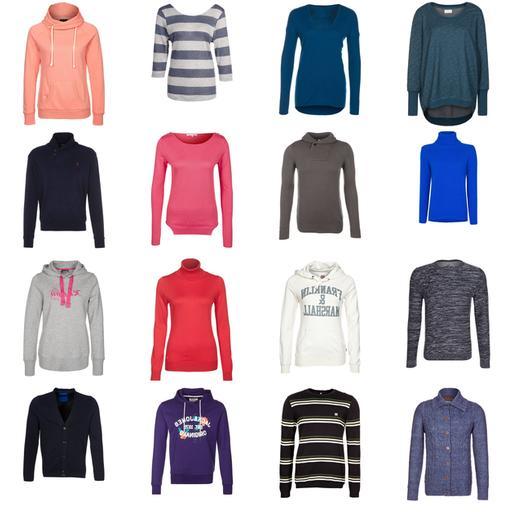}} }

\subfigure[$\{x_1^j\}_{j=1}^{16}$ generated from a fixed human image $x_1$ and the article set $\{y_i\}_{i=1}^{16}$. The image $x_1$ is in the top left corner of (a).]{\fbox{\includegraphics[width=6.9cm]{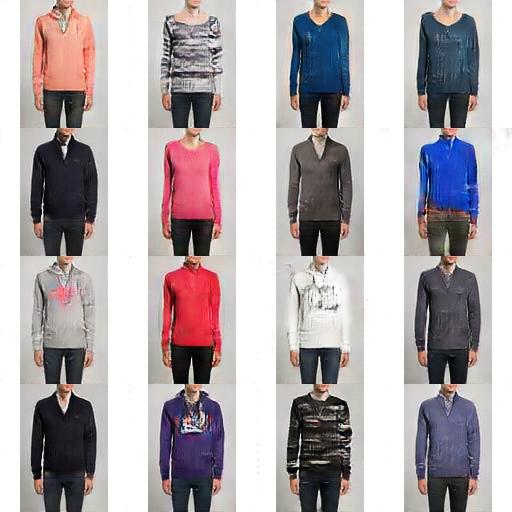}} }\hspace{5mm}
\subfigure[$\{x_i^1\}_{i=1}^{16}$ generated from a fixed article image $y_1$ and different people images $\{x_i\}_{i=1}^{16}$. The article image $y_1$ is in the top left corner of (b).]{\fbox{\includegraphics[width=6.9cm]{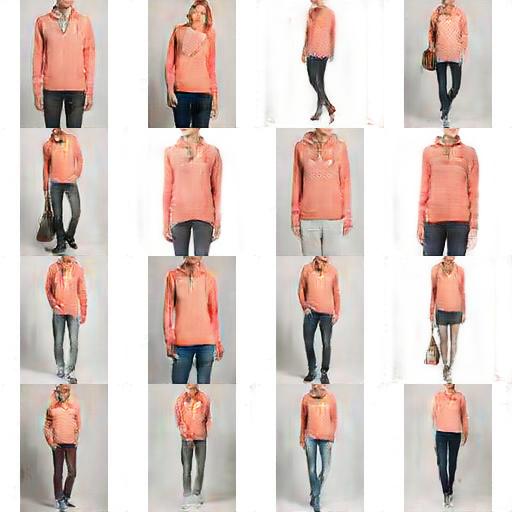}} }
\caption{Combining different human models (a) and articles (b) demonstrates the generalization ability of CAGAN, as shown in the plots (c) and (d), where generated images combining people and clothes in novel ways is demonstrated. Result images are of size 128x96 pixel.}
\label{fig_many}
\end{figure*}

Figure \ref{fig_many}(c) shows how we can swap different clothes on the same human, or apply the same fashion article on different humans (d). This also shows clearly that our model generalizes well and can combine any (coming from the training distribution) human and article in a visually appealing way.
As a sidenote, we could obtain a similar performance using images $x,y$ not contained in the original training set $\{x_i,y_i\}$, as long as the photoshoot style is consistent with the training data -- the CAGAN does not memorize the training data.
Note, however, that the model is good at transferring the colors and rough structures of clothes but not the fine textures, see Figure \ref{fig_many}(c), top row 2nd image (from left to right) and bottom row 3rd image. 


\section{Discussion}
\subsection{Related methods}
Another recent method for generating people with clothes~\cite{clothnet} uses segmented data from the Chictopia \cite{chictopia} dataset, and applies auto-encoders and image translation cGAN (as in \cite{pix2pix}) to generate images of humans using segmented body regions as conditioning images. However, they lack the ability to specify what piece of clothing to generate exactly -- e.g. they generate any upper body garment on the specified  segmentation mask region, but cannot control how it looks exactly. In addition, the requirement for semantic segmentation data can be a limitation in practice: fashion companies do not usually gather such data. Inferring and labeling such data can lead to extra costs: per-pixel segmentation is expensive.

  In contrast, using available fashion images -- as CAGAN does -- is a more natural fit to the fashion domain: such data is already available in huge quantities, and also allows to precisely specify what article to paint.

\subsection{Future work}
We showed the concept and first results of a novel type of GAN that can swap clothes on humans and offer new possibilities to image manipulation for fashion purposes.
We plan to continue this work and improve upon it both in model architecture and application. 

We want to use more data and try more exciting fashion article swapping scenarios. Being able to change all fashion item categories (upper/lower body garments, shoes, accessories) on a human picture is essential for the full ``virtual try-on" experience.
We will also examine in detail how much good segmentations of human model images can improve the overall results. Having at least a foreground-background segmentation can indeed be beneficial, see~\cite{texturegan}. We can easily augment CAGAN with segmentation masks if they are provided for the human model pictures, either by directly overwriting the mask $\alpha$ or by using them as a prior. 

We plan to improve the neural network architecture in several ways. (i) Examine other color spaces, e.g. \textbf{Lab} instead of \textbf{RGB}.
(ii) Test whether using texture descriptors (as in \cite{texturegan}) can improve the results when swapping clothing with specific textile patterns -- currently the CAGAN is inaccurate with complex textures.
(iii) Analyze whether embedding of the conditioning information (the articles $y$) can lead to better flow of information in the neural network and faster convergence, as in \cite{reed2016generative}. Right now the generator needs both to analyze the visual descriptors and localize the article image $y_i$ on the human image $x_i$. Having an embedding of the visual description of the article may help the earlier layers to focus on the correct region where the old fashion article is located.



{\small
\bibliographystyle{ieee}
\bibliography{bibi}
}

\end{document}